# Enhancing Skin Lesion Diagnosis with Ensemble Learning

Xiaoyi Liu*
Ira A. Fulton Schools of Engineering
Arizona State University
Tempe, USA
*Corresponding author: xliu472@asu.edu

Lianghao Tan
W. P. Carey School of Business
Arizona State University
Tempe, USA
ltan22@asu.edu

Ge Shi
Independent Researcher
Tempe, USA
geshi@asu.edu

Zhou Yu
Department of Mathematics, Statistics, and Computer Science
University of Illinois at Chicago
Chicago, USA
zyu33@uic.edu

Yafeng Yan
Department of Computer Science
Stevens Institute of Technology
Hoboken, USA
yanyafeng0105@gmail.com

*Abstract*— Skin lesions are an increasingly significant medical concern, varying widely in severity from benign to cancerous. Accurate diagnosis is essential for ensuring timely and appropriate treatment. This study examines the implementation of deep learning methods to assist in the diagnosis of skin lesions using the HAM10000 dataset, which contains seven distinct types of lesions. First, we evaluated three pre-trained models: MobileNetV2, ResNet18, and VGG11, achieving accuracies of 0.798, 0.802, and 0.805, respectively. To further enhance classification accuracy, we developed ensemble models employing max voting, average voting, and stacking, resulting in accuracies of 0.803, 0.82, and 0.83. Building on the best-performing ensemble learning model, stacking, we developed our proposed model, SkinNet, which incorporates a customized architecture and fine-tuning, achieving an accuracy of 0.867 and an AUC of 0.96. This substantial improvement over individual models demonstrates the effectiveness of ensemble learning in improving skin lesion classification.

I. INTRODUCTION

Skin lesions are a common medical concern with an increasing number of patients each year. These lesions vary widely in severity; some, like melanoma, are highly cancerous, while others, such as vascular lesions, are benign and pose no harm [1]. Additionally, certain lesions, such as actinic keratoses, can develop into skin cancer if left untreated [1]. Accurately detecting and classifying different types of skin lesions is critical to ensure appropriate and timely treatment.

In this paper, we apply deep learning to assist in skin lesion diagnosis using the HAM10000 dataset, which includes seven types of lesions, ranging from benign to cancerous. We tested three pre-trained models—MobileNetV2, ResNet18, and VGG11—achieving accuracies of 0.798, 0.802, and 0.805, respectively. To improve accuracy, we developed ensemble models using max voting, average voting, and stacking, which increased accuracies to 0.82 and 0.83, except for max voting, which achieved 0.803. Building on this, our proposed SkinNet model, with customized architecture and fine-tuning, achieved an accuracy of 0.867 and an AUC of 0.96, significantly outperforming individual models. This demonstrates the potential of ensemble learning in enhancing skin lesion classification.

II. BACKGROUND

*A. Machine Learning*

Machine learning (ML) plays a crucial role in today's world, with numerous use cases such as user credit risk prediction [2], payment security systems [3], automated robotic pathfinding [4], energy consumption prediction [5], obsessive-compulsive disorder prediction [6], VR experience prediction [7], and data protection [8].

*B. Deep Learning*

Deep learning (DL) methods, a branch of ML, have delivered remarkable results across various industries. Numerous applications leverage DL technology, including financial risk behavior prediction [9], financial risk management [10], cryptocurrency analytics [11], heart rate prediction [12], stock prediction [13], and healthcare [14]. This paper will focus on DL in computer vision (CV), with applications such as pose-invariant face recognition [15], 3D scene reconstruction [16], super-resolution image reconstruction [17], brain tumor detection and segmentation [18] [19], and more.

*C. Skin Lesions*

The dataset comprises 7 distinct types of skin lesions categorized into skin cancers, lesions with the potential to develop into skin cancer, and benign lesions with no harmful effects. Among the skin cancers, melanoma is the most aggressive type, while basal cell carcinoma is the most prevalent and less aggressive form [1]. Actinic keratoses are precancerous lesions that have the potential to develop into skin cancer [1]. For benign lesions with no threat to health, the dataset includes melanocytic nevi (commonly known as moles), benign keratosis-like lesions, dermatofibromas, and various vascular lesions [1]. All types of skin abnormalities can be seen in Figure 1.

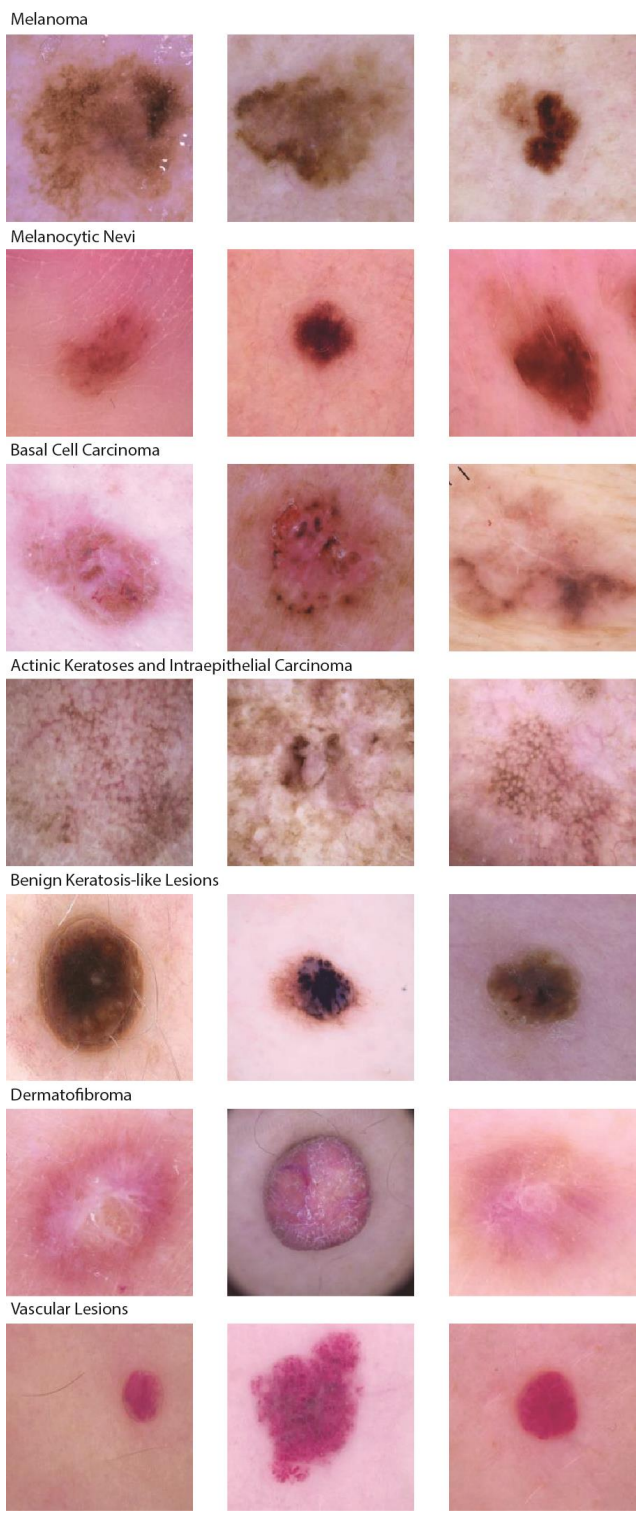

Figure 1. Skin Disease Types

### D. Pre-trained Models

*1) VGG11:* VGG11 is made up of eleven layers, incorporating both convolutional and fully connected layers [20]. VGG11 has been extensively applied in various domains; however, it contains a substantial amount of parameters. The exact number of parameters for different models can be found in Table I.

*2) ResNet-18:* ResNet introduces residual learning components that allow the model to skip certain layers, aiding in alleviating the vanishing gradient issue. This enables the model to go deeper and supports the creation of more complex architectures [21].

*3) MobileNetV2:* The MobileNet architecture, with its small number of parameters, is specifically optimized for mobile and other computationally limited systems. MobileNetV2's inverted residuals and linear bottlenecks further enhance the model's efficiency [22].

TABLE I. PARAMETER COMPARISON

| Model | Number of Parameters |
|---|---|
| VGG11 | 132.9 million |
| ResNet18 | 11.7 million |
| MobileNetV2 | 3.4 million |

### III. METHODOLOGY

#### A. Dataset

The publicly accessible Skin Cancer MNIST: HAM10000 dataset [23] is utilized in this research. It comprises 10,015 dermatoscopic images featuring 7 distinct categories of skin abnormalities. Initially, the dataset was cleaned by removing duplicate images from the same patient ID, leaving only one image per patient. This step will remove nearly identical images from the dataset. After cleaning, a total of 5,973 images remained. Table II provides further details on the image count for each skin abnormality type.

TABLE II. NUMBER OF IMAGES FOR EACH SKIN LESION CATEGORY

| Num. of Images | Description |
|---|---|
| 491 | Melanoma (mel) |
| 4322 | Melanocytic nevi (nv) |
| 261 | Basal cell carcinoma (bcc) |
| 182 | Actinic keratoses and intraepithelial carcinoma (akiec) |
| 581 | Benign keratosis-like lesions (bkl) |
| 58 | Dermatofibroma (df) |
| 78 | Vascular lesions (vasc) |

#### B. Image Augmentation and Preprocessing

TABLE III.  MODEL COMPARISON

| Model | Accuracy | F1 Score | Recall | Precision | AUC |
|---|---|---|---|---|---|
| ResNet18 | 0.8024 | 0.7732 | 0.8024 | 0.7756 | 0.9260 |
| MobileNetV2 | 0.7984 | 0.7794 | 0.7984 | 0.7697 | 0.9257 |
| VGG11 | 0.8051 | 0.7846 | 0.8051 | 0.7743 | 0.9184 |
| **Ensemble Learning: (ResNet18 + MobileNetV2 + VGG11)** | **Accuracy** | **F1 Score** | **Recall** | **Precision** | **AUC** |
| Max Voting | 0.8037 | 0.7882 | 0.8037 | 0.7839 | 0.9333 |
| Average Voting | 0.8224 | 0.8033 | 0.8224 | 0.7941 | 0.9466 |
| Stacking | 0.8344 | 0.8202 | 0.8344 | 0.8119 | 0.9474 |
| Stacking-FT | 0.8598 | 0.8593 | 0.8598 | 0.8612 | 0.9600 |
| **SkinNet (Ours)** | **0.8678** | **0.8644** | **0.8678** | **0.8633** | **0.9609** |

The images in the HAM10000 dataset originally had a resolution of 450 x 600 pixels. However, since the skin lesion information is generally centered in the image, the images are first cropped to 450 x 450 pixels to retain all relevant information without altering the image's aspect ratio. After cropping, the images are resized to 224 x 224 pixels.

Image augmentation and preprocessing are applied during training. The images are flipped horizontally and vertically with a 50% probability, followed by a random rotation of up to 10 degrees. Next, the images are resized and converted to tensors. Lastly, all images undergo normalization.

*C. Learning Rate*

In this study, the initial learning rate is set to 0.01, paired with a momentum value of 0.9. The optimization process is carried out using the Stochastic Gradient Descent (SGD) optimizer. The learning rate is scaled down by 0.1 after each set of ten epochs. This allows the model to optimize its learning by taking smaller, more accurate steps in weight adjustments, reducing the risk of overshooting the optimal solution. If there is no improvement in validation accuracy for 10 consecutive epochs, the training will be stopped.

*D. Pre-trained Models*

MobileNetV2, ResNet18, and VGG11 were selected for their proven effectiveness in image classification tasks, achieving accuracies of 0.798, 0.802, and 0.805, respectively. To adapt these models to our dataset, which includes 7 classes, the final fully connected layers were modified accordingly. In MobileNetV2, the classifier was replaced with a new layer that includes a dropout layer with a 20% chance of dropping out, followed by a fully connected layer with 7 outputs. Similarly, the final layers of ResNet18 and VGG11 were adjusted to accommodate the 7 output categories.

*E. Ensemble Learning*

Ensemble learning is a powerful machine learning technique that combines multiple models to achieve better outcomes than relying on a single model. In this paper, we will analyze two voting ensemble methods: max voting and average voting, as well as stacking ensemble learning. In the max voting approach, each model (ResNet18, VGG11, and MobileNetV2) processes the input image and generates a set of logits, denoted as $z_1$, $z_2$, and $z_3$, respectively. Each logit vector z contains a score for each class. The ensemble method then computes the element-wise maximum of these logits across the three models [24].

$$z_{max} = \max(z_1, z_2, z_3) \quad (1)$$

Here, $z_{max}$ is the final logit vector produced by the Max Voting Ensemble. The final predicted class is determined by using the final logit $z_{max,i}$ to identify the class with the highest score [24].

In the average voting approach, each model (ResNet18, VGG11, and MobileNetV2) processes the input image and generates a set of logits, denoted as $z_1$, $z_2$, and $z_3$, respectively. Each logit vector z contains a score for each class. The ensemble method then computes the element-wise average of these logits across the three models [24].

$$z_{avg} = (z_1 + z_2 + z_3) / 3 \quad (2)$$

Here, $z_{avg}$ is the final logit vector produced by the Average Voting Ensemble. The final predicted class is determined by using the final logit $z_{avg,i}$ to identify the class with the highest score [24].

In the stacking ensemble learning approach, predictions from multiple models are combined using a meta-learner. The fully connected layer in the meta-learner learns from the outputs of each model (ResNet18, VGG11, and MobileNetV2) and optimally combines these predictions to enhance overall performance [24]. Unlike conventional stacking ensemble methods, in this paper, the outputs from the individual models are logits rather than probabilities.

Max voting, average voting, and stacking ensemble learning attained accuracies of 0.803, 0.822, and 0.834, in that order. Further studies will focus on stacking ensemble learning as it achieves the best performance.

*F. Proposed Deep Learning Models*

The stacking ensemble model combining ResNet18, VGG11, and MobileNetV2 will first be fine-tuned by unfreezing the last feature layers in MobileNetV2 and ResNet18 while keeping VGG11's layers frozen. Unfreezing VGG11 would reduce efficiency with minimal gain, while MobileNetV2 and ResNet18 offer performance improvements with far fewer parameters. The fine-tuned model, referred to as Stacking-FT, achieved an accuracy of 0.859 and an AUC of 0.96, which is a significant improvement over the original stacking ensemble model, which had an accuracy of 0.834 and an AUC of 0.947.

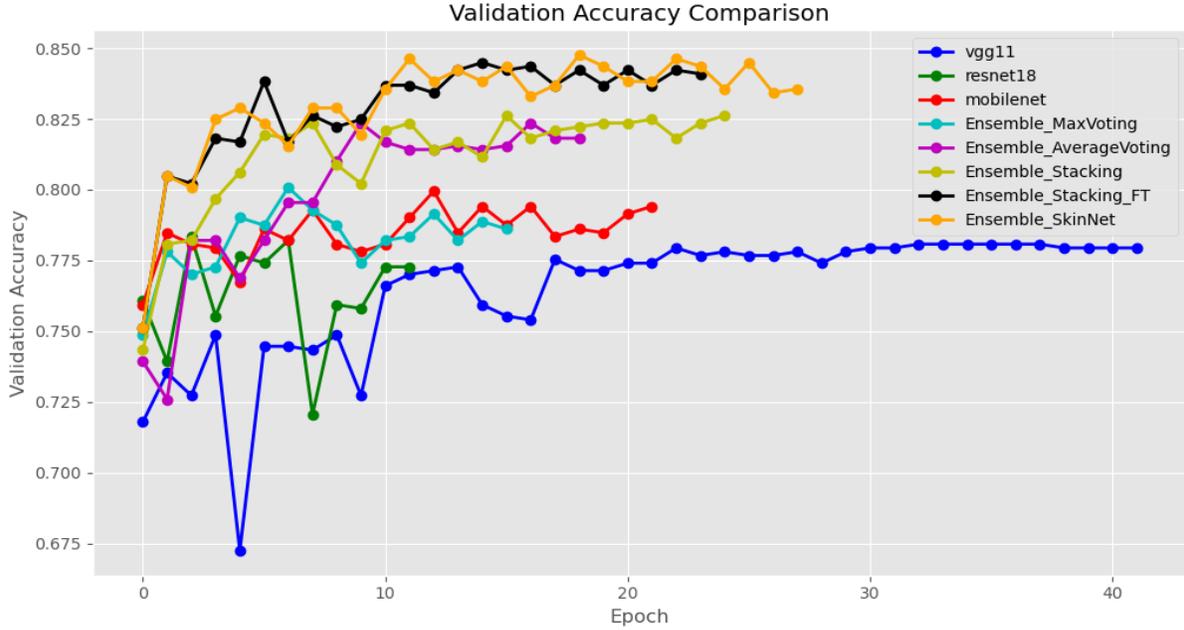

Figure 2. Validation Accuracy Comparison

Our proposed model, referred to as SkinNet, builds upon the fine-tuned stacking ensemble model (Stacking-FT). In this enhanced version, the stacking ensemble combines the outputs of MobileNetV2, ResNet18, and VGG11. Given that more layers are unfrozen in MobileNetV2 and ResNet18, a weighted approach is introduced in the meta-learner to appropriately balance the contributions of each model. The outputs from MobilenetV2 and resnet18 are multiplied by a weight factor of 1.2 to emphasize their contributions while keeping the weight the same for VGG11.

Mathematically, let $z_1$, $z_2$, and $z_3$ represent the logits from MobileNetV2, ResNet18, and VGG11, respectively. The weighted logits are calculated as follows:

$$z_{weighted} = \text{concat}(1.2 \cdot z_1, 1.2 \cdot z_2, z_3) \quad (3)$$

These weighted logits are then concatenated and passed through a fully connected layer that takes the combined output from the three models and produces the final classification. Our proposed model, SkinNet, achieved the accuracy and AUC of 0.867 and 0.960 which further improved from Stacking-FT. More details on SkinNet's architecture are in Figure 3.

### G. Performance Measurements

Accuracy and F1-score are used in this classification task, along with precision and recall as metrics. In addition to these, AUC is also employed as a metric due to the imbalance in the dataset. As shown in Table II, some classes have over 4,300 images, while others have only 58 images. AUC measures the ability of a model to distinguish between classes, making it particularly useful for evaluating performance on imbalanced datasets. The cross-entropy loss function is employed to evaluate the model's performance by comparing the predicted probabilities with the true class labels.

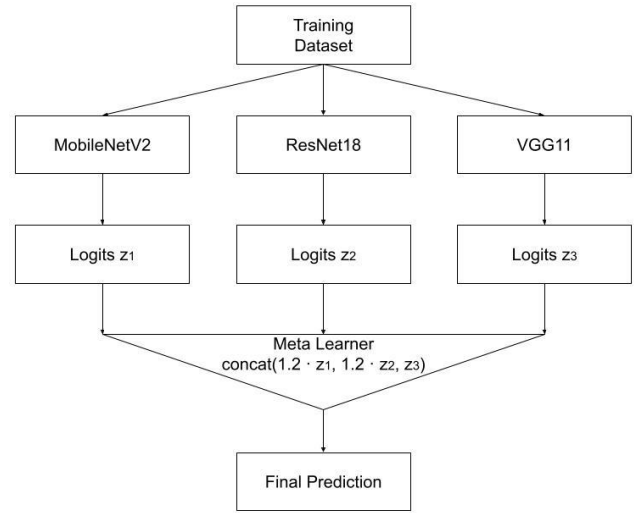

Figure 3. SkinNet Architecture

### IV. EVALUATION AND DISCUSSION OF RESULTS

From Figure 2, the pre-trained models perform the worst in terms of validation accuracy. The stacking ensemble learning outperformed the other two ensemble methods: max voting and average voting. The proposed SkinNet model achieved the best performance among all models, with the stacking ensemble learning with fine-tuning ranking a close second. In terms of training epochs, VGG11 required over 40 epochs, while the proposed SkinNet converged in less than 30 epochs.

Regarding the performance of the testing dataset, as shown in Table III, all three ensemble learning models (max voting, average voting, and stacking) outperformed the individual models, except max voting, which did not surpass the individual models in accuracy. The stacking ensemble learning with fine-tuning further improved performance on the specific task of skin lesion classification.

The proposed model, SkinNet, achieved the highest results across all metrics, with an accuracy of 0.867, an F1 score of 0.863, and an AUC of 0.96. This represents a significant improvement over any single pre-trained model.

## V. CONCLUSION

For the assignment of classifying skin lesions, three pre-trained models (MobileNetV2, ResNet18, and VGG11) were tested on the HAM10000 dataset, achieving accuracies of 0.798, 0.802, and 0.805, respectively. Three ensemble learning models combining these pre-trained models (max voting, average voting, and stacking) were then tested, resulting in improved accuracies of 0.82 and 0.83, except for max voting, which achieved an accuracy of 0.803. Building on stacking ensemble learning, we developed our proposed model, SkinNet, which incorporates customized architecture and fine-tuning. SkinNet achieved an accuracy of 0.867 and an AUC of 0.96, representing a significant improvement over any individual model. Ensemble learning has demonstrated its effectiveness by combining the strengths of multiple models, leading to superior outcomes in medical image classification tasks.